# A MODIFICATION OF PARTICLE SWARM OPTIMIZATION USING RANDOM WALK

Rajesh Misra  
Lecturer at S.A.Jaipuria College  
rajeshmisra.85@gmail.com

Kumar S. Ray  
Professor at Indian Statistical Institute  
ksray@isical.ac.in

Abstract -Particle swarm optimization comes under lot of changes after James Kennedy and Russell Eberhart first proposes the idea in 1995. The changes has been done mainly on Inertia parameters in velocity updating equation so that the convergence rate will be higher. We are proposing a novel approach where particle's movement will not be depend on its velocity rather it will be decided by constrained biased random walk of particles. In random walk every particles movement based on two significant parameters, one is random process like toss of a coin and other is how much displacement a particle should have. In our approach we exploit this idea by performing a biased random operation and based on the outcome of that random operation, PSO particles choose the direction of the path and move non-uniformly into the solution space. This constrained, non-uniform movement helps the random walking particle to converge quicker then classical PSO. In our constrained biased random walking approach, we no longer needed velocity term ($V_i$), rather we introduce a new parameter (K) which is a probabilistic function. No global best particle ($P_{Gbest}$), local best particle ($P_{Lbest}$), Constriction parameter (W) are required rather we use a new term called $P_{targ}$ which is loosely influenced by $P_{Gbest}$. We test our algorithm on five different benchmark functions, and also compare its performance with classical PSO and Quantum Particle Swarm Optimization (QPSO). This new approach have been shown significantly better than basic PSO and sometime outperform QPSO in terms of convergence, search space, number of iterations.

## I. Introduction

In Intelligent search and Optimization algorithms Particle Swarm Optimization is one of the strongest candidate. The idea of this algorithm generally originates from the social behavior of birds flocking, fish schooling and animal hoarding. This group of animals interact each other's for finding food in a collaborative manner. PSO is highly influenced by the logic of how they interact, how they quickly move from one place to another in a huge search area. Two major component of PSO algorithm is particles position and velocity. Based on social behavior particles velocity changed and applied to their initial position to get new position. Classical PSO method does not require any gradient information straightway mathematical operation helps to converge the particle and it is quite easy to understand.[11][12]. This basic method sometimes trap into local minima and unable to converge into solutions which encourage lot of researchers to optimize the algorithm in such a way that it can quickly converge and can successfully avoid local minima issue. Lot of variation over PSO has been applied over the past decade such velocity clamping method[8], changes in inertia Weight [10], introduction of constriction parameter[16]. Every approach has their own merits and demerits. Such as in velocity clamping method sometime particle will not converge in local area rather stuck in global optima, Constriction coefficient method sometime cause unnecessary fluctuation of particles.

Our approach in this paper is completely different from previously described approaches. We apply Random walk in PSO particle movement. As per our knowledge not much research work has been done in this are so far other that [2][3].Random Walk is an mathematical object moving in a mathematical space influenced by random, stochastic process. We find a handful of similarity in PSO movement with Random movement of particle. PSO particle will move based on their calculated velocity which can be replaced by random stochastic movement of particle based on statistical function. In classical PSO particle follow a global best particle and their motion carefully controlled by neighbor particles speed; it can also be replaced by a constrained based biased random walk which already has application in network science field. Novelty of our approach is elimination of 1) Velocity term and 2) unlike using two parameters as 'global best' and 'local best' we simply introduce a single parameter as 'target particle' and 3) apply biased random walk on graph strategy by creating PSO graph.

This paper is organized as follows: section II discusses basic PSO algorithm, section III discusses simple random walk, and biased random walk. Section IV cover detail approach of our proposed modification of PSO algorithm. Section V gives experimental setup and section VI explains experimental results and explore conclusion.

## II. Basic PSO algorithm

In 1995 James Kennedy and Russell Eberhart proposed an evolutionary algorithm that create a ripple in Bio-inspired algorithmic approach called Particle Swarm Optimization (PSO). In a simple term it is a method of optimization for continuous non-linear function. As this method influenced by swarming theory form biological world like fish schooling, bird swarming etc. In every PSO method there is two important terms, one is position and another is velocity which will be updated by the following formula. Let $X_i(t)$ denote position of particle i in the search space at time t. Then the position will be calculated as follows:

$$X_i(t + 1) = X_i(t) + V_i(t + 1) \quad \text{--(1)}$$

Where $V_i(t+1)$ is the velocity of particle i at time (t+1), which will be computed based on this following formula:

$$V_i(t) = V_i(t-1) + C_1 * R_1(P_{LB}(t) - X_i(t-1)) + C_2 * R_2(P_{GB}(t) - X_i(t-1)) \quad \text{---- (2)}$$

Where

C1, C2 = Constants determine the relative influence on social and cognitive components, also known as learning rate, often set to same value to give each component equal weights.

R1, R2 = random values associated with learning rate components to give more robustness.

PLB = Particle Local Best position – it is the historically best position of the ith particle achieved so far

PGB = Particle Global Best position – it is the historically best position of the entire swarm, basically position of a particle which achieve closest solution.

Equation (2) is Kennedy and Eberhart's original idea, after that lot of different research has been going on, based on that one of the remarkable idea comes up by Shi and Eberhart [14] of addition of a new factor called "inertia weight" or "w ". After addition of inertia weight the eq. (2) becomes as follows –

$$V_i(t) = w * V_i(t-1) + C_1 * R_1 (P_{LB}(t) - X_i(t-1)) + C_2 * R_2 (P_{GB}(t) - X_i(t-1)) \quad --- (3)$$

This inertia weight helps to balance local and global search abilities, small weight means local search and larger weight means global search.

### PSO algorithm

Pseudo code of the basic PSO algorithm as follows –

```
For each particle
   Initialize particle
END

Do
   For each particle
      Calculate fitness value
      If the fitness value is better than the best fitness value (P_LB) in history
         set current value as the new P_LB
   End

   Choose the particle with the best fitness value of all the particles as the P_GB
   For each particle
      Calculate particle velocity according equation (2)
      Update particle position according equation (1)
   End
While maximum iterations or minimum error criteria is not attained
```

### III. Random walk

#### a. Simple Random Walk

The term random walk was first introduced by Karl Pearson in 1905.[21]. It is a mathematical formalization of a path which consist of successive random steps .The basic idea of random walk is quite simple.  Let's consider one Dimensional Random walk on an integer line Z. If a person starts his journey from 0 and at each step he moves his left (+1) or right (-1) with equal probability, then the question arises can that person ever come back to his starting location 0?

Mathematically in One Dimensional Random walk, if we consider independent random variables $Z_1, Z_2,\ldots$ where each variable is either 1 or -1 with a 50% probability of either value and Let $S_0 = 0$ then $S_n = \sum_{j=1}^{n} Z_j$. **The Expectation $E(S_n)$** of $S_n$ is zero. Which implied that [1] that expected translation distance after n steps is $\sqrt{n}$. In answer to the previous question yes, after n number of random steps that person will roaming around location 0.

Random walk has significant number of application in many different fields like computer science, Ecology, Economics, Chemistry, and Physics. There are various different types of random walks like Random walk on Graphs, Self-interacting random walk, Biased Random walk in graphs. Different types of model is random walk is also exists such as Lattice Random Walk which is simple random walk on a regular lattice, one dimensional random walk, higher dimensional random walk, and Gaussian Random Walk. Let explain algorithmically Simple Random walk as follows -

**Procedure SRandomWalk**

Step 1: Compute a random function e.g. tossing a fair coin.
Step 2: if HEAD the move Left
Step 3 : Else move Right.
   This above procedure is uniform movement where random process is also unbiased.

  b. **Biased Random walk**

In simple random walk particles movement is decided by equal probability, as a result after N steps particle is close to the position where it started. But in biased random walk the jump of the particle from the current state to the next state is influenced by unequal probability. As for example let consider a biased coin tossed and 70% chances coming Head and 30% Tail. Again if Head means left movement and Tail means Right movement than it is quite obvious that particle moves left more than Right. So after N number of steps we will find particles expected position more left direction than right.

Numerous research work has been done on Network Science where Biased Random Walk on Graphs applied for structural analysis of Network Flow Graph [5], Social Network [4] or Ad-Hoc network [6]. There have been many different representation of biased random walk. A general one which is used in undirected graph by J. Gómez-Gardeñes; V. Latora[9] for calculating the entropy rate to characterize a diffusion process on a complex network is as follows –

Let's imagine an undirected graph where each node has an attribute $α_i$. Now if there is a walker who is traversing the graph is currently on node j and wishes to jump node i, will calculate the probability of jumping form node j to i. The probability will be $P_{ij}{}^α = \frac{α_i A_{ij}}{\sum_k α_k A_{kj}}$, where $A_{ij}$ indicates as a weight parameter or edge value form node j to node i. Depending on the application, α value going to be different in each case. It is a influencing factor on the edge connecting two nodes, in social network graph it could be "attractor value" of a person, or it could be some differentiable inherent characteristic of individual nodes. Based on probabilistic outcome the Walker will decide which node i he has to jump.

**Procedure BiasedRandomWalk**

Step 1: Compute biased Random process like tossing a unfair coin.
Step 2: if HEAD move left with uniform displacement ( like +1 or -1)
Step 3: Else Right with uniform displacement ( like +1 or -1)

This procedure perform a unfair coin tossing which leads to either most of the time HEAD or TAIL, as a result the particle will move left or right more. So the particle will not roaming near to where it started, it will actually move from one place to another. But at each step the movement is uniform.

**Procedure ConstrainedBiasedRandomWalk**

Step 1: Compute biased Random process like tossing a unfair coin.
Step 2: if HEAD move left with non-uniform displacement (like +Δ or - Δ).
Step 3: Else Right non-uniform displacement (like +Δ or - Δ).

This procedure is same as earlier one but here in each step, movement is not uniform, a Δ amount of displacement is added or subtracted every time the particle is moving.

## IV. Proposed PSO Method Modified By Biased Random Walk

A. General Idea –

Our proposed approach begins with the construction of PSO graph, a dynamic, undirected, and weighted, complete graph which we will discuss in detail in next section. After the PSO graph constructed attribute values $α_i$ will be computed for each particle. Calculation $α_i$ will be later explained. Next, for each iteration each particle will compute $A_{ij}$ which is weight parameter. Once $A_{ij}$ is calculated for every particle, a probability function $P_{ij}^α = \frac{α_i A_{ij}}{\sum_k α_k A_{kj}}$, will be computed for every edges connected to all other particles. A random number r is picked after that, if r < lowest probability ($P_{ij}^α$) among all the path or edges from node j then the particle connected to the other end of the lowest probability edge will be choose as possible target particle $P_{targ}$, This $P_{targ}$ is a new parameter concept introduced here, Else highest probability particle will be choose as $P_{targ}$. Our intended particle $P_j$ will jump to the nearest location of $P_{targ}$. Algorithmically we can summarize this as follows –

**Procedure RandomWalkPSO**

Step 1: Construct PSO graph.
Step 2: Compute attribute values $α_i$ of each node.
Step 3: For each node j.
Step 4: Compute $A_{ij}$ weight parameter for each edge form node j to all other nodes i.

Step 5: Calculate probability $P_{ij}{}^\alpha = \frac{\alpha_i A_{ij}}{\sum_k \alpha_k A_{kj}}$, for all edges connecting to node j.

Step 6: $\theta_{max}$ – maximum probability value and $\theta_{min}$ – minimum probability value for node j among all probability values calculated in step 5.

Step 7: Pick a random number r in range 0 to 1.

Step 8: if r < $\theta_{min}$; choose the particle holding $\theta_{min}$ probability as $P_{targ}$.

Step 9: Else choose the particle holding $\theta_{max}$ probability as $P_{targ}$.

Step 10: Compute $K_x$ and $K_y$ using equation (5) based on newly calculated $P_{targ}$ and update new position of the particle $P_j$

Step 11: End for

Step 12: Repeat from Step 2 to Step 11 until maximum iteration or minimum error criteria is attained.

### B. PSO Graph Construction(G(P,E))

In PSO Graph G (P,E) each particles will be considered as node $p_i \in P$ and the Euclidian distance from $p_j$ to $p_i$ will be edge $e_j \in E$. This graph is dynamic in the way that it will change its structure for each iteration. In each iteration the edges will change because the distance between nodes change. Each node or particle is connected to all other nodes forming a mesh structure. As it is undirected graph so the distance from node $p_i$ to node $p_j$ is same as node $p_j$ to $p_i$ for $i^{th}$ iteration. This graph is also considered as weighted graph and the weight is computed as Euclidian distance between $p_j(x_i,y_i)$ to $p_j(x_j,y_j)$ as follows –

Euclidian distance $(p_j, p_i) = \sqrt{(x_j - x_i)^2 + (y_j - y_i)^2}$

In the PSO graph we use the term 'PSO particle' and 'PSO node' interchangeably as they represent same object in both PSO search space and PSO graph.

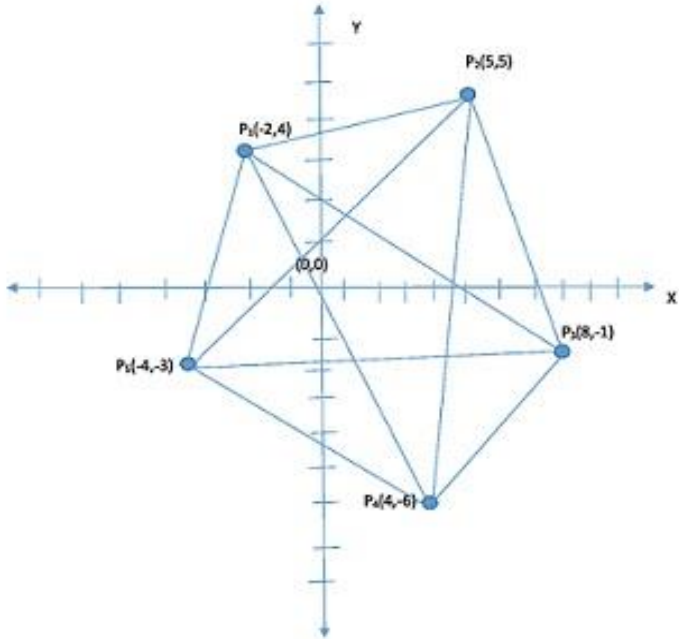

Figure 1 : a Random PSO graph with 5 nodes.
In the above example graph a snapshot of particle distribution has been shown and also how the graph from that snapshot is constructed also shown. In the above 2 dimensional coordinated system let's imagine there is 5 particle $P_1(-2,4), P_2(5,5), P_3(8,-1), P_4(4,-6), P_5(-4,-3)$ distributed over the search space. We connect each particle with other particle as edge. We compute distance and mark them as weight in each edge.

### C. Attribute Computation ($α_i$)

We need to compute attribute value for each nodes in PSO graph. This will be used in probability computation in later phase. Computation of attribute is rank based, here a integer vector $Rank(1:N) = \{1,2,3,4…N\}$ where N is the swarm size, is maintained . First we will compute fitness of all the particle in the swarm, then we assign maximum rank which is 'N' in this case, form the Rank Vector to the particle which achieve highest fitness, and then the next highest will get next rank N-1 and 3rd highest will be assigned N-2 so on and the particle which is furthest from the desired solution will be assigned minimum rank 1. We continue this process until all the particles assigned with their respective rank value which we called $α_i$ here. We need to remember here that if the application is on maximization problem then highest fitness mean the particle which achieve maximum fitness value and if it is minimization problem then highest-fitness-particle means the particle which achieve minimum fitness value among others. Algorithmically we can summarize the attribute computation is as follows –

**Procedure AttributeComputation ($α_i$)**

Step 1: Compute fitness value of all particles.
Step 2: Identify the particle which achieved highest fitness value for the maximization problem or lowest fitness value for the minimization problem.
Step 3: Assign maximum rank which is N in our case because we consider rank vector as a vector of 1 to N where N is swarm size.
Step 4: Assign next maximum rank to the next highest fitness particle and so on.
Step 5: Repeat step 3 to step 4 until all particle got their rank.

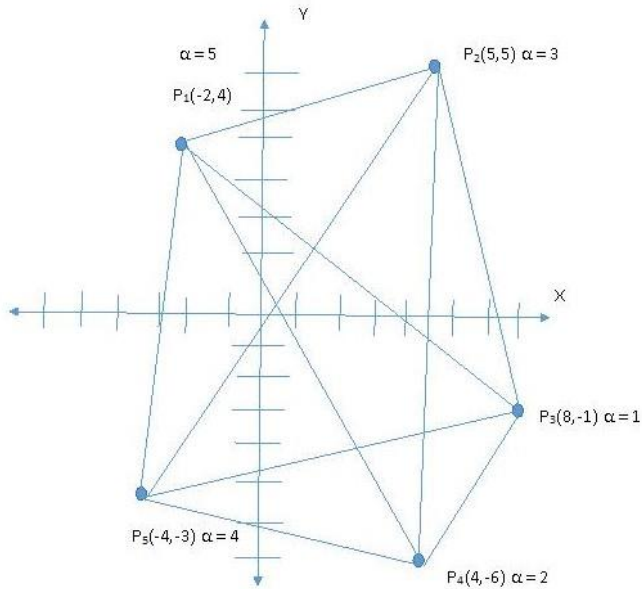

Figure – 2: same above example graph with nodes calculated with attribute value α.
In the above example figure we consider a minimization problem where the closest particle towards the (0,0) coordinate is the highest-fitness-particle. Accordingly $P_1$ gets highest rank $a=5$ $P_5$ gets $a=4$, $P_2$ gets $a=3$, $P_4$ gets $a=2$, $P_3$ gets $a=1$.

### Explanation of Rank value assignment:
In Particle Swarm Optimization, particles in each iteration try to converge in solution space, that's why they try to come close to that particle which is near to the solution space as much as possible. Keeping that in our mind, we assign maximum rank value i.e. N to that particle which is closest to the solution so far. We did this because in our probability computation the rank value which is $α_i$ play an important role. It will be multiplied with $A_{ij}$ which is weight parameter and this ($α_i$ x $A_{ij}$) will decide which particle will be chosen as $P_{targ}$. In our case $A_{ij}$ is distance value, now for closest particle $A_{ij}$ is 1 and $α_i$ is maximum and multiplication between this two will give least value among all. So in worst case particle will never diverge too far, rather it will choose nearest position as much as possible.

   D.  Weight Parameter Calculation($A_{ij}$)

We already discuss the weight parameter calculation in PSO graph construction section. Each edge carries a weight which is the distance from the source node to the destination node. Suppose we have to calculate the weight of each edge from the particle $P_j$. Then we will compute the Euclidian distance from node $P_j$ to $P_i$ where i =1 to N and i≠j, and assign that value as a weight of the edge joining $P_j$, $P_i$. We will call this value $A_{ij}$. For $P_j$ to $P_j$ the $A_{ij}$ = 1 which is non-zero minimal integer value.

### Procedure WeightParamCal ($A_{ij}$)

Step 1: For all node $P_i$ ($x_i,y_i$) from node $P_j(x_j,y_j)$

Step 2: Compute $A_{ij} = \sqrt{(x_j - x_i)^2 + (y_j - y_i)^2}$

Step 3: End For

Step 4 : For $P_i$ to $P_i$ ; $A_{ii}$ = 1, a non zero constant value.

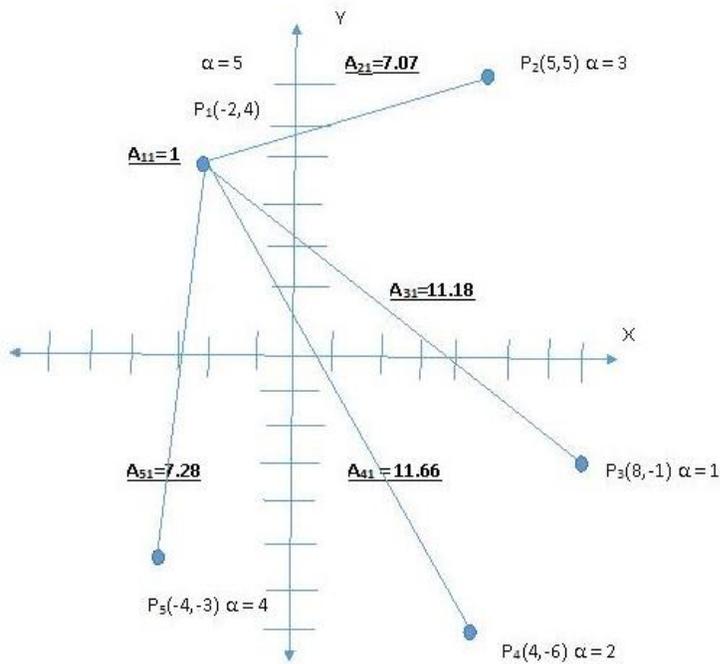

Figure – 3: PSO graph after calculation of weight parameters $A_{ij}$

Form the above figure we can see from particle $P_1$ all the other distance to the particle $P_2,P_3,P_4,P_5$ are calculated as follows –
$A_{21}$ = Euclidian Distance from $P_1$ to $P_2$ = 7.07
$A_{31}$ = Euclidian Distance from $P_1$ to $P_3$ = 11.18
$A_{41}$ = Euclidian Distance from $P_1$ to $P_4$ = 11.66

$A_{51}$ = Euclidian Distance from $P_1$ to $P_5$ = 7.28
And
$A_{11}$ = Euclidian Distance from $P_1$ to $P_1$ = 1.

E. Probability Computation ($P_{ij}{}^\alpha$)

Now the probability computation for each possible path from node $P_j$ to $P_i$ will be computed. This probability will guide us where will be the next position of the particle $P_j$. Each edge emanating from $P_j$ must contain $P_{ij}{}^\alpha$. Probability computation function from node j to node i is as follows –

$$P_{ij}{}^\alpha = \frac{\alpha_i A_{ij}}{\sum_k \alpha_k A_{kj}} \qquad -- \qquad (4)$$

Where
$\alpha_i$ - is the attribute value of all the particle in the PSO graph. And
$A_{ij}$ – is the weight parameter for the edge from node j to node i.
Equation (4) is a simple probability distribution function based on weight on the edges of a graph. Earlier this probability function is utilized in Entropy rate diffusion process by J. Gómez-Gardeñes; V. Latora [7].We borrowed the idea of this probability function in our PSO graph which we can effectively apply. This probability function can be used where biased random walk can be applied in a graph, weather the graph is network graph, social graph or our PSO graph.

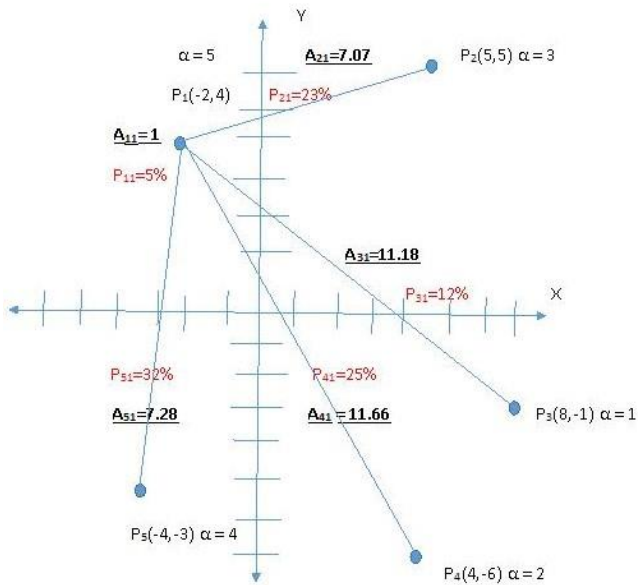

Figure – 4: PSO graph with probability calculation.

The above figure shows the same PSO graph after probability calculation. Once this calculation done, each path from $P_1$ we got a probability which basically indicates the statistical chances of choosing that path as well as that particle as its target particle $P_{targ}$.

$P_{11} = 5/(1 \times 11.18 + 2 \times 11.66 + 3 \times 7.07 + 4 \times 7.28 + 5 \times 1) = 5/89.83 = 5\%$
$P_{21} = (3 \times 7.07) / 89.83 = 23\%$
$P_{31} = (1 \times 11.18) / 89.83 = 12\%$
$P_{41} = (2 \times 11.66) / 89.83 = 25\%$
$P_{51} = (4 \times 7.28) / 89.83 = 32\%$

According to the probability calculation highest probability particle is $P_5$. And lowest is $P_1$.

### F. Target Particle Selection ($P_{targ}$)

Rather calculating local best ($P_{Lbest}$) and global best particle ($P_{Gbest}$) here we consider only single particle called as target particle, we are eliminating global best or local best particle by introducing the concept of target particle ($P_{targ}$). But we are not completely devoid of the concept of global and local best. Global best particle indicates the particle which is more close to the solution, we use this concept in our attribute calculation ($\alpha_i$) where the more close -to-the-solution particle got highest rank onwards and during probability calculation ($P_{ij}$) the product term ($\alpha_i \times A_{ij}$) got highest probability values which indicate which particle to choose. So global best particle $P_{Gbest}$ became $P_{targ}$ here.

But after finding $P_{targ}$ a particle ($P_i$) cannot just jump to the position of $P_{targ}$ because of randomness. In random walk theory a random process like coin toss, rolling dice or picking a random card will decide which path the particle should choose. In our case we pick a random number and that will decide whether the $P_{targ}$ will be chosen or some other particle.

It is quite clear that, suppose the closest particle is $P_k$ so it $A_{ik}$ will be 1, and $\alpha_k = N$, so after probability calculation this particles probability will be lowest. If r is less than this probability then $P_k$'s target particle will be $P_k$ itself that means it will not change its position as it is in its best position so far. Due to this in our approach particles are not unnecessarily jumping randomly.

Even if r is greter than least probability value, still the particle will not jump too high due to this ($\alpha_i \times A_{ij}$) product term.

In basic PSO, there is a concept of local best particle $P_{Lbest}$ which store the last best position of the particle $P_i$. That $P_{Lbest}$ restrict a sudden jump towards $P_{Gbest}$ because particle's movement also influenced by its neighbor position and speed. In our approach we are eliminating $P_{Lbest}$, because we don't need to store earlier best position, it will not help to take decision about the new position of the particle in our algorithm.

Once we know which path to choose we will move nearest position of that $P_{targ}$ particle but the amount of movement is decided by two variable $K_x$ and $K_y$ which we will discuss in next section. In random walk theory this $K_x$ and $K_y$ is considered as constraints and has different values in different iteration. In a way our particle movement is not uniform rather following a Constrained Biased Random Walk.

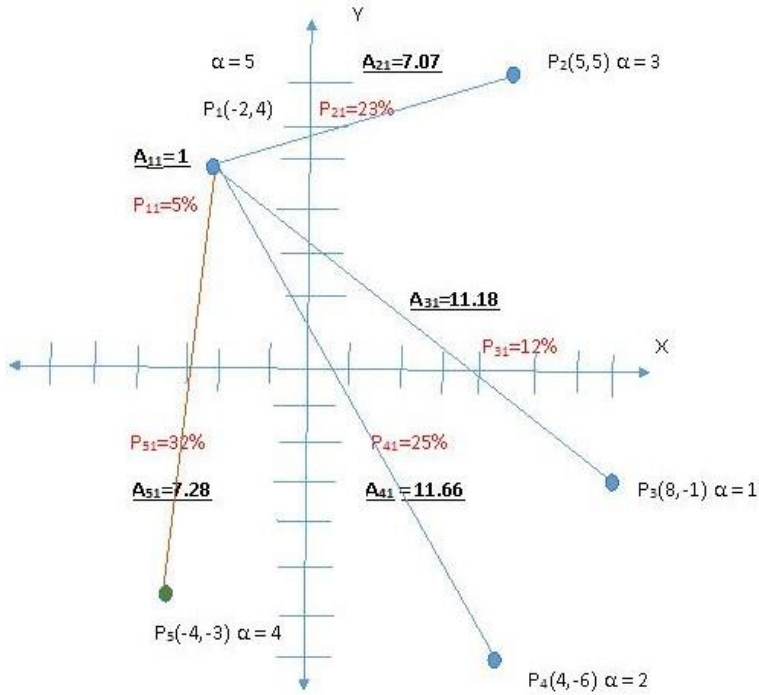

Figure 5: the green color node is selected as $P_{targ}$ and $P_1$ particle will move to that position.

G. New Position Computation($P_{new}$)

In random walk after selecting the direction and target particle if the particle moves unit values (+1) or (-1) then it will call uniform random walk. But in our approach we choose non-uniform amount of movement by incorporating two variable $K_x$ and $K_Y$ In this constraint random walk $K_x$ denotes the amount of displacement from the initial position in X- direction and similarly $K_Y$ indicates same displacement in Y-direction.

Calculation of $K_x$ and $K_Y$

During biased random walk once $P_{targ}$ get selected our next step will be to find out how much displacement delta ($\Delta$) we need to add in initial position of the particle to reach $P_{targ}$ after n steps. Let's consider $n\Delta$ is the increasing or positive steps and $n(1-\Delta)$ is the decreasing or negative steps. So in order to get the value of $\Delta$ to maximize the probability of reaching target after n steps is –

$$P_{targ} - P_i = n(1-\Delta) - n\Delta$$

$$\text{So } \Delta = \frac{1 - \frac{P_{targ} - P_i}{n}}{2} \quad \text{---- (5)}$$

This X- direction delta ($\Delta$) will be called $K_x$ and Y- Direction delta ($\Delta$) will be called $K_Y$.

Gaussian distribution parameter

$K_x$ and $K_Y$ are two variables that help a particle not to trap in local minima or maxima but this two parameters are not only parameters. We also introduce another parameters called $G_{parm}$ which restrict particle for not jumping to high. $G_{parm}$ is a normal continuous probability distribution function which is calculated as follows –

$$G_{parm}(x \mid \mu, \sigma^2) = \frac{1}{\sqrt{2\pi\sigma^2}} e^{-\frac{(x-\mu)^2}{2\sigma^2}} \quad \text{for X- direction} \quad \text{--- (6)}$$

$$G_{parm}(y \mid \mu, \sigma^2) = \frac{1}{\sqrt{2\pi\sigma^2}} e^{-\frac{(y-\mu)^2}{2\sigma^2}} \quad \text{for Y- direction} \quad \text{--- (7)}$$

Here,
μ - is the mean or expectation of the distribution.
σ is the standard deviation. And
σ² is the variance.

Final New Position Computation

After calculating equation (5) and Equation (6) and (7) we can now compute the probable position of the particle as follows –

$$P_{new}(x) = P_i(x) + K_x + G_{parm}(x \mid \mu, \sigma^2) \text{ for X – direction} \quad \text{---(8)}$$
$$P_{new}(y) = P_i(y) + K_y + G_{parm}(y \mid \mu, \sigma^2) \text{ for Y – direction} \quad \text{---(9)}$$

This new position is addition of 3 different terms we discussed above where $P_i$ is the initial position of the particle and $K_x/K_y$ is displacement parameters in X- and Y- direction and $G_{parm}$ is Gaussian distribution parameters.

H. The Parameters we eliminated from basic PSO in our modified PSO -

a. **Computation of Velocity (v) of particle is no longer needed.**

> Velocity term in basic PSO reflects that fact that how much a particle should move from its initial position. In our approach, during implementing biased random walk we calculate $K_x$ and $K_y$ which are two probabilistic values that influence the amount of movement of particle in an iteration. We no longer needed to calculate velocity term. We should keep in mind that velocity term $V_x$ and $V_y$ are important pillar of basic PSO but they require a longer computation involving $P_{Lbest}$, $P_{Gbest}$, $C_1$, $C_2$, $r_1, r_2$. Our $K_x$ and $K_y$ are much simpler probabilistic function without usage of the position of local and global best particle and lot random variables. It is less computation heavy and much quicker in operation.

b. **Introduction of unified target particle $P_{targ}$ instead of two separate concept of $P_{Gbest}$ and $P_{Lbest}$.**

During discussion of Target Particle Selection section we discuss the reason behind $P_{targ}$. We can say loosely $P_{Gbest}$ is $P_{targ}$. But we removed the idea of $P_{Lbest}$ as it does not suit any purpose in our random walk approach.

## V. Experimental Setting

a. Benchmark functions -

In our experiment we use 3 well known single objective optimization function. Those are Sphere Function, Rosenbrock Function, Rastrigin function, and 2 multi objective optimization function those are Test function- 4 [8] and Schaffer function N-1[9]. Here the proposed algorithm is compared with basic Particle Swarm Optimization algorithm (PSO) and also its latest variants Quantum Particle Swarm Optimization (QPSO). In Table 1 we show the mathematical representation of those function used.

Table 1: Functions with mathematical representation with Global minimum values.

| Function number | Function Name | Mathematical Representation | Global minimum |
|---|---|---|---|
| $F_1$ | Sphere Function | $f(x) = \sum_{i=1}^{n} x_i^2$ | $f(x_1.....x_n) = f(0...0)=0$ |
| $F_2$ | Rosenbrock Function | $f(x) = \sum_{i=1}^{n-1} [100(x_{i+1} - x_i^2)^2 + (x_i - 1)^2]$ | Min = n=2 →f(1,1)=0<br>n=3<br>→f(1,1,1)= 0<br>n>3<br>→f(1...1)=0 |
| $F_3$ | Rastrigin function | $f(x) = An + \sum_{i=1}^{n}[x_i^2 - A\cos(2\pi x_i)]$ | $f(0,0) = 0$ |
| $F_7$ | Test function- 4 (multi objective) | $Minimize = \begin{cases} f_1(x,y) = x^2 - y \\ f_2(x,y) = -0.5x - y - 1 \end{cases}$ | -7 < x,y <4 |

| F$_8$ | Schaffer function N-1(multi objective) | $Minimize = \begin{cases} f_1(x) = x^2 \\ f_2(x) = (x-2)^2 \end{cases}$ | -A <= x <= A; A is from 10 to 10$^5$ |

For the purpose of comparison, the asymmetric initialization method has been used [10].

    b. Population size -

As per Shi and Eberhart [14] population size does not play a major role in performance analysis of PSO algorithm. Van den Bergh and Engelbrecht [11] shows that swarm size indeed play a significant role as solution quality will improve slightly by increasing the swarm size. In our work we keep population size 20, 40, 80, 160 as used by Shi and Eberhart [14].

In our experiment we use Intel i3 processor with 3.00GHz clock speed, 4.00GB RAM and 64 bit Windows Operating system. The programming language we used is Matlab 2015a.

## VI. Experimental Results And Discussion

In table 2 to table 4 we show single objective optimization function performance of our proposed algorithm and also compare basic PSO and QPSO performance. And in table 5 and 6 we show same performance for multiobjective functions.

Table 2: performance comparison for Sphere Function

| Function | Population size | Dimension | Basic PSO | QPSO | RWPSO | Mean Best Fitness | | |
|---|---|---|---|---|---|---|---|---|
| F$_1$ | | | Iterations | Iterations | Iterations | Basic PSO | QPSO | RWPSO |
| | 20 | 10 | 7 | 5 | 56 | 0.000 | 0.000 | 0.000 |
| | | 20 | 19 | 12 | 78 | | | |
| | | 30 | 21 | 22 | 104 | | | |
| | 40 | 10 | 31 | 19 | 103 | 0.000 | 0.000 | 0.000 |
| | | 20 | 43 | 23 | 99 | | | |
| | | 30 | 65 | 39 | 102 | | | |
| | 80 | 10 | 78 | 67 | 103 | 0.000 | 0.000 | 0.000 |
| | | 20 | 91 | 78 | 134 | | | |
| | | 30 | 123 | 89 | 149 | | | |
| | 160 | 10 | 203 | 91 | 233 | 0.000 | 0.000 | 0.000 |

| | | 20 | 251 | 123 | 245 | | | |
| | | 30 | 323 | 176 | 267 | | | |

Table 3: performance comparison for Rosenbrock Function

| Function | Population size | Dimension | Basic PSO | QPSO | RWPSO | Mean Best Fitness | | |
|---|---|---|---|---|---|---|---|---|
| $F_1$ | | | Iterations | Iterations | Iterations | Basic PSO | QPSO | RWPSO |
| | 20 | 10 | 12 | 10 | 28 | 3.115 | 1.214 | 1.100 |
| | | 20 | 17 | 10 | 39 | 6.121 | 4.123 | 3.189 |
| | | 30 | 25 | 14 | 40 | 18.124 | 15.121 | 17.156 |
| | 40 | 10 | 30 | 23 | 56 | 4.121 | 0.121 | 0.221 |
| | | 20 | 41 | 20 | 96 | 34.345 | 23.901 | 19.125 |
| | | 30 | 62 | 35 | 102 | 67.891 | 45.112 | 32.112 |
| | 80 | 10 | 70 | 63 | 108 | 3.674 | 1.673 | 1.918 |
| | | 20 | 90 | 73 | 132 | 67.890 | 45.190 | 34.901 |
| | | 30 | 111 | 91 | 156 | 123.421 | 120.101 | 101.192 |
| | 160 | 10 | 156 | 89 | 155 | 3.788 | 1.319 | 1.215 |
| | | 20 | 245 | 134 | 165 | 89.991 | 67.914 | 35.910 |
| | | 30 | 310 | 186 | 167 | 145.911 | 131.956 | 94.011 |

Table 4: performance comparison for Rastrigin function.

| Function | Population size | Dimension | Basic PSO | QPSO | RWPSO | Mean Best Fitness | | |
|---|---|---|---|---|---|---|---|---|
| $F_1$ | | | Iterations | Iterations | Iterations | Basic PSO | QPSO | RWPSO |
| | 20 | 10 | 19 | 13 | 16 | 91.257 | 67.901 | 45.100 |
| | | 20 | 78 | 67 | 78 | 214.989 | 201.561 | 167.911 |
| | | 30 | 81 | 69 | 84 | 319.901 | 300.781 | 230.101 |

| | 40 | 10 | 42 | 35 | 233 | 70.124 | 70.091 | 64.091 |
| | | 20 | 67 | 59 | 63 | 180.982 | 180.901 | 170.981 |
| | | 30 | 91 | 78 | 105 | 299.781 | 290.891 | 256.901 |
| | 80 | 10 | 156 | 90 | 123 | 36.125 | 23.100 | 17.067 |
| | | 20 | 304 | 177 | 154 | 87.891 | 67.901 | 45.901 |
| | | 30 | 471 | 203 | 149 | 208.981 | 200.901 | 189.451 |
| | 160 | 10 | 671 | 342 | 313 | 24.891 | 22.901 | 20.901 |
| | | 20 | 891 | 370 | 345 | 72.190 | 67.912 | 52.051 |
| | | 30 | 978 | 671 | 467 | 131.145 | 112.901 | 90.001 |

Table 5: performance comparison for Test function- 4

| Function | Population size | Dimension | Basic PSO | QPSO | RWPSO | Mean Best Fitness | | |
|---|---|---|---|---|---|---|---|---|
| $F_1$ | | | Iterations | Iterations | Iterations | Basic PSO | QPSO | RWPSO |
| | 20 | 10 | 70 | 59 | 56 | -2.1219 | -0.141 | -3.2100 |
| | | 20 | 119 | 110 | 79 | -6.9012 | -1.892 | -3.1891 |
| | | 30 | 223 | 220 | 156 | 3.9012 | 2.189 | 0.1902 |
| | 40 | 10 | 31 | 19 | 103 | -4.0102 | -0.192 | 0.2210 |
| | | 20 | 43 | 23 | 99 | -5.1902 | -3.190 | -1.0292 |
| | | 30 | 65 | 39 | 102 | -1.9022 | -0.901 | 0.1011 |
| | 80 | 10 | 78 | 67 | 103 | -2.2563 | -1.127 | 0.0120 |
| | | 20 | 91 | 78 | 134 | 1.9091 | 3.102 | 0.9012 |
| | | 30 | 123 | 89 | 149 | 6.9021 | 3.901 | 0.9101 |
| | 160 | 10 | 203 | 91 | 233 | 0.6721 | 0.901 | 0.7816 |
| | | 20 | 251 | 123 | 245 | 0.9021 | 0.961 | 1.9012 |
| | | 30 | 323 | 176 | 267 | 0.9561 | 1.902 | 1.0312 |

Table 9: performance comparison for Schaffer function N-1

| Functi | Populat | Dimension | Basic | QPSO | RWPSO | Mean Best Fitness |
|---|---|---|---|---|---|---|

| on | ion size | | PSO | | | | | |
|---|---|---|---|---|---|---|---|---|
| $F_1$ | | | Iterations | Iterations | Iterations | Basic PSO | QPSO | RWPSO |
| | 20 | 10 | 82 | 59 | 61 | $1.212 \times 10^4$ | $0.2341 \times 10^3$ | $0.1210 \times 10^4$ |
| | | 20 | 99 | 93 | 78 | $0.9112 \times 10^4$ | $0.6732 \times 10^4$ | $0.9212 \times 10^4$ |
| | | 30 | 24 | 130 | 110 | $1.9236 \times 10^4$ | $1.023 \times 10^3$ | $1.0123 \times 10^4$ |
| | 40 | 10 | 81 | 45 | 54 | $3.1902 \times 10^5$ | $3.1092 \times 10^4$ | $2.1093 \times 10^5$ |
| | | 20 | 243 | 204 | 139 | $-4.192 \times 10^5$ | $-2.343 \times 10^4$ | $-1.283 \times 10^4$ |
| | | 30 | 361 | 259 | 255 | $9.1092 \times 10^5$ | $3.1023 \times 10^4$ | $1.7845 \times 10^4$ |
| | 80 | 10 | 742 | 633 | 433 | $5.892 \times 10^5$ | $3.1029 \times 10^5$ | $1.2931 \times 10^5$ |
| | | 20 | 911 | 781 | 614 | $3.2901 \times 10^5$ | $2.8945 \times 10^5$ | $1.9742 \times 10^5$ |
| | | 30 | 1363 | 1094 | 1203 | $3.2903 \times 10^5$ | $2.9023 \times 10^5$ | $1.7683 \times 10^5$ |
| | 160 | 10 | 301 | 290 | 133 | $1.2901 \times 10^4$ | $1.0465 \times 10^4$ | $1.6775 \times 10^4$ |
| | | 20 | 1453 | 1341 | 1192 | $-2.394 \times 10^5$ | $-2.347 \times 10^5$ | $-2.31 \times 10^5$ |
| | | 30 | 1891 | 1301 | 1232 | $1.9032 \times 10^4$ | $1.2031 \times 10^4$ | $0.9021 \times 10^5$ |

All the experimental results shown on those tables contain iterations and Mean Fitness values. Iterations means to achieve that mean best fitness how much iteration needed by each of 3 different PSO methods. The values shown in the iterations column are taken average after 50 runs. The mean best values which we show here achieved by the PSO methods are considered 80% of total population size. The C1 and $C_2$ values are considered as 2 and 2 and random number range considered as 0.9 to 0.4. This is same as used by Shi and Eberhart [14].

VII. Performance Graph Comparison For 4 Different Fitness Functions

Figure -6 to 9 shows performance comparison for 2 single objective and 2 multiobjective fitness function.

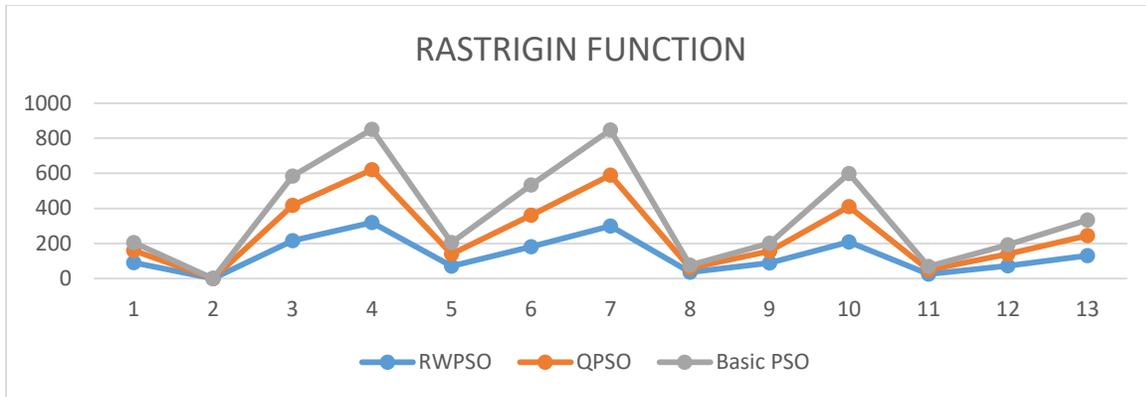

Figure – 6: Performance comparison for Rastrigin Function. Vertical dimension shows fitness values for this minimization function.

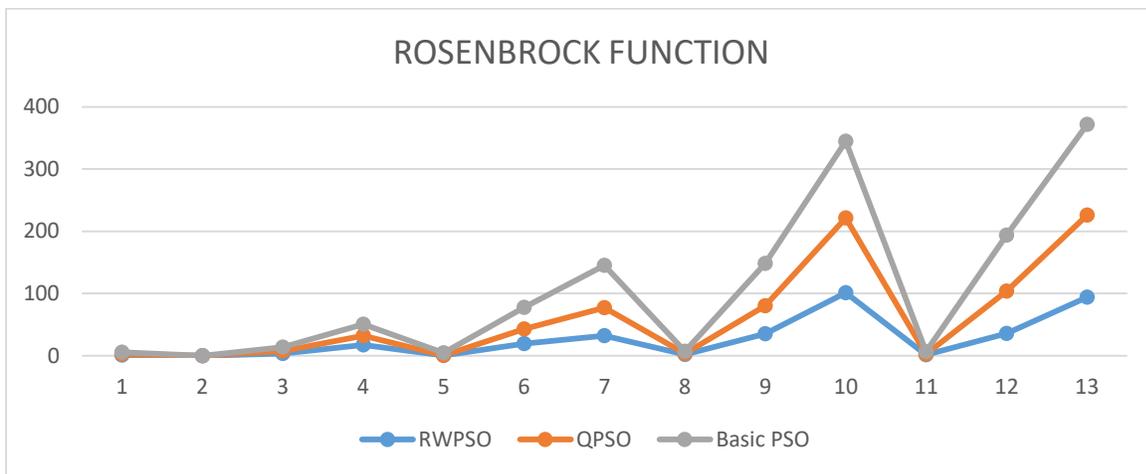

Figure – 7: Performance comparison for Rosenbrock Function. Vertical dimension shows fitness values for this minimization function

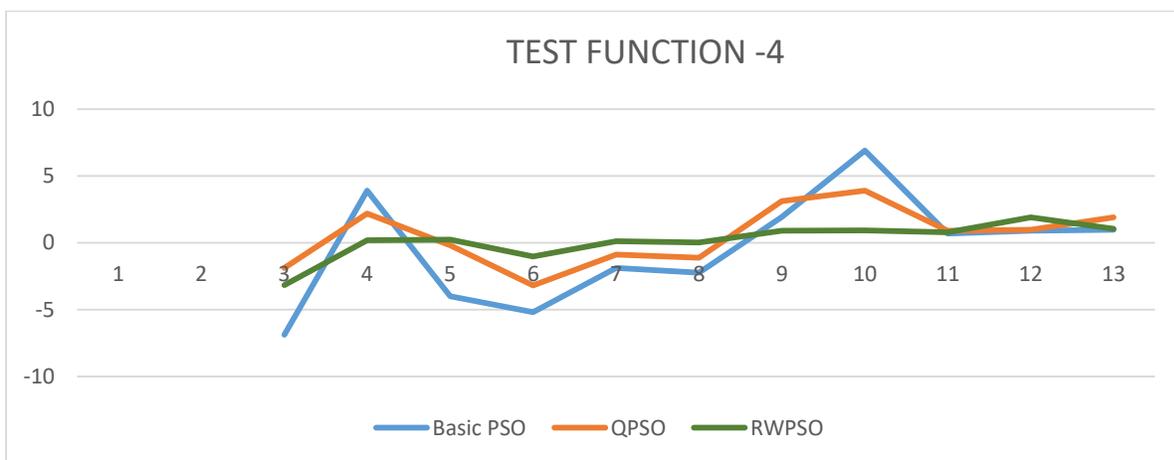

Figure – 8: Performance comparison for TEST Function -4 multiobjective function.

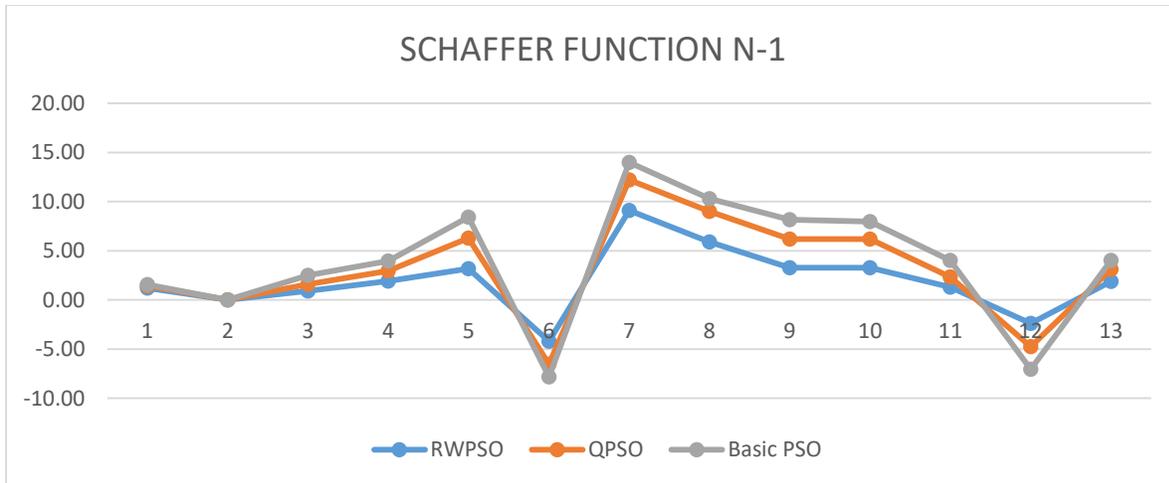

Figure -9: Performance comparison for Schaffer N-1 Function.

## Conclusion

In this paper we followed extensive experiment on different fitness function which shows our proposed method works quite well in terms of performance, convergence and iterations. With increasing population size and dimension the Random Walk PSO perform better than basic PSO and sometimes exceed the performance of Quantum PSO as well. Some of the fields we can do better like $G_{parm}$, instead of using simple normal distribution we can use Rayleigh distribution which is quite applicable in higher dimension random walk.